\documentclass[letterpaper,dvipsnames]{article} 
\usepackage{aaai2026}  
\nocopyright
\usepackage{times}  
\usepackage{helvet}  
\usepackage{courier}  
\usepackage[hyphens]{url}  
\usepackage{graphicx} 
\def\UrlFont{\rm}  
\usepackage{natbib}  
\usepackage{caption} 
\usepackage{algorithm}
\usepackage{algorithmic}

\usepackage{newfloat}
\usepackage{listings}

\usepackage{booktabs}
\usepackage{multirow}
\usepackage{amssymb}  
\usepackage{pifont}   
\usepackage{amsmath}
\usepackage{array}
\usepackage{diagbox}
\usepackage{arydshln}
\usepackage{siunitx}
\usepackage{bibentry}
\usepackage[most]{tcolorbox} 
\usepackage{subcaption}
\usepackage{mdframed}
\usepackage{tabularx}
\usepackage{lipsum}
\usepackage{placeins}

\usepackage{hyperref}

\hypersetup{
    pdftitle={CoCoNUTS: Concentrating on Content while Neglecting Uninformative Textual Styles for AI-Generated Peer Review Detection},
    pdfauthor={Yihan Chen, Jiawei Chen, Guozhao Mo, Xuanang Chen, Ben He, Xianpei Han, Le Sun},
    colorlinks=true,
    linkcolor=blue,
    citecolor=blue,
    filecolor=magenta,
    urlcolor=blue
}

\DeclareCaptionStyle{ruled}{labelfont=normalfont,labelsep=colon,strut=off} 
\floatstyle{ruled}
\newfloat{listing}{tb}{lst}{}
\floatname{listing}{Listing}
\title{CoCoNUTS: Concentrating on Content while Neglecting Uninformative \\ Textual Styles for AI-Generated Peer Review Detection}
\author{
    Yihan Chen\textsuperscript{\rm 1,\rm 2},
    Jiawei Chen\textsuperscript{\rm 1,\rm 2},
    Guozhao Mo\textsuperscript{\rm 1,\rm 2},
    Xuanang Chen\textsuperscript{\rm 1},
    Ben He\textsuperscript{\rm 1,\rm 2},
    Xianpei Han\textsuperscript{\rm 1},
    Le Sun\textsuperscript{\rm 1}
}
\affiliations{
    \textsuperscript{\rm 1}Chinese Information Processing Laboratory, Institute of Software, Chinese Academy of Sciences\\
    \textsuperscript{\rm 2}University of Chinese Academy of Sciences\\
    \{chenyihan20241,chenjiawei2024,moguozhao2024,chenxuanang,xianpei,sunle\}@iscas.ac.cn\\
    benhe@ucas.ac.cn
}

\mdfdefinestyle{ExampleBox}{%
    linecolor=black,       
    outerlinewidth=0.5pt,  
    roundcorner=2pt,       
    innertopmargin=\topskip, 
    innerbottommargin=\topskip, 
    innerrightmargin=10pt, 
    innerleftmargin=10pt,  
    frametitlefont=\bfseries, 
    backgroundcolor=white, 
}

\newtcolorbox{myprompt}[2][]{
  enhanced,
  breakable, 
  colback=black!5, 
  colframe=black!60, 
  fonttitle=\bfseries, 
  title={#2}, 
  attach boxed title to top left={xshift=0.5cm, yshift=-2mm},
  boxed title style={
    colback=black!60,
  },
  verbatim
}

\begin{document}

\maketitle

\begin{abstract}

The growing integration of large language models (LLMs) into the peer review process presents potential risks to the fairness and reliability of scholarly evaluation. 
While LLMs offer valuable assistance for reviewers with language refinement, there is growing concern over their use to generate substantive review content.
Existing general AI-generated text detectors are vulnerable to paraphrasing attacks and struggle to distinguish between surface language refinement and substantial content generation, suggesting that they primarily rely on stylistic cues. When applied to peer review, this limitation can result in unfairly suspecting reviews with permissible AI-assisted language enhancement, while failing to catch deceptively humanized AI-generated reviews.
To address this, we propose a paradigm shift from style-based to content-based detection.
Specifically, we introduce CoCoNUTS, a content-oriented benchmark built upon a fine-grained dataset of AI-generated peer reviews, covering six distinct modes of human-AI collaboration. 
Furthermore, we develop CoCoDet, an AI review detector via a multi-task learning framework, designed to achieve more accurate and robust detection of AI involvement in review content.
Our work offers a practical foundation for evaluating the use of LLMs in peer review, and contributes to the development of more precise, equitable, and reliable detection methods for real-world scholarly applications. Our code and data will be publicly available at https://github.com/Y1hanChen/COCONUTS.
\end{abstract}
\section{Introduction}

Peer review is an essential component of academic publications. However, the rapid advancement of Large Language Models (LLMs) and the increasing workload of reviewers have raised significant concerns about the misuse of LLMs in the peer review process. These concerns are reflected in official policies of academic conferences. For instance, ACL's policy permits using LLMs for auxiliary tasks like language polishing but strictly prohibit using them for generating the substantive content. 
Although these policies aim to regulate the use of LLMs, a study indicates a rising trend in the use of LLMs for the substantive content modification of reviews in some conferences~\cite{liang2024monitoring}.  
This misuse not only creates a risk of data leakage but also results in reviews of unreliable quality. Researches indicates that LLMs struggle to fairly evaluate scientific contributions~\cite{ye2024we}, often generating reviews without details and actionable suggestions~\cite{zhou-etal-2024-llm,du-etal-2024-llmsassistnlp}, and are vulnerable to manipulation, making them unsuitable for autonomous peer review. 
Therefore, detecting the extent of AI involvement in peer reviews is essential. 

\begin{figure*}[t]
    \centering
    \includegraphics[width=0.97\linewidth]{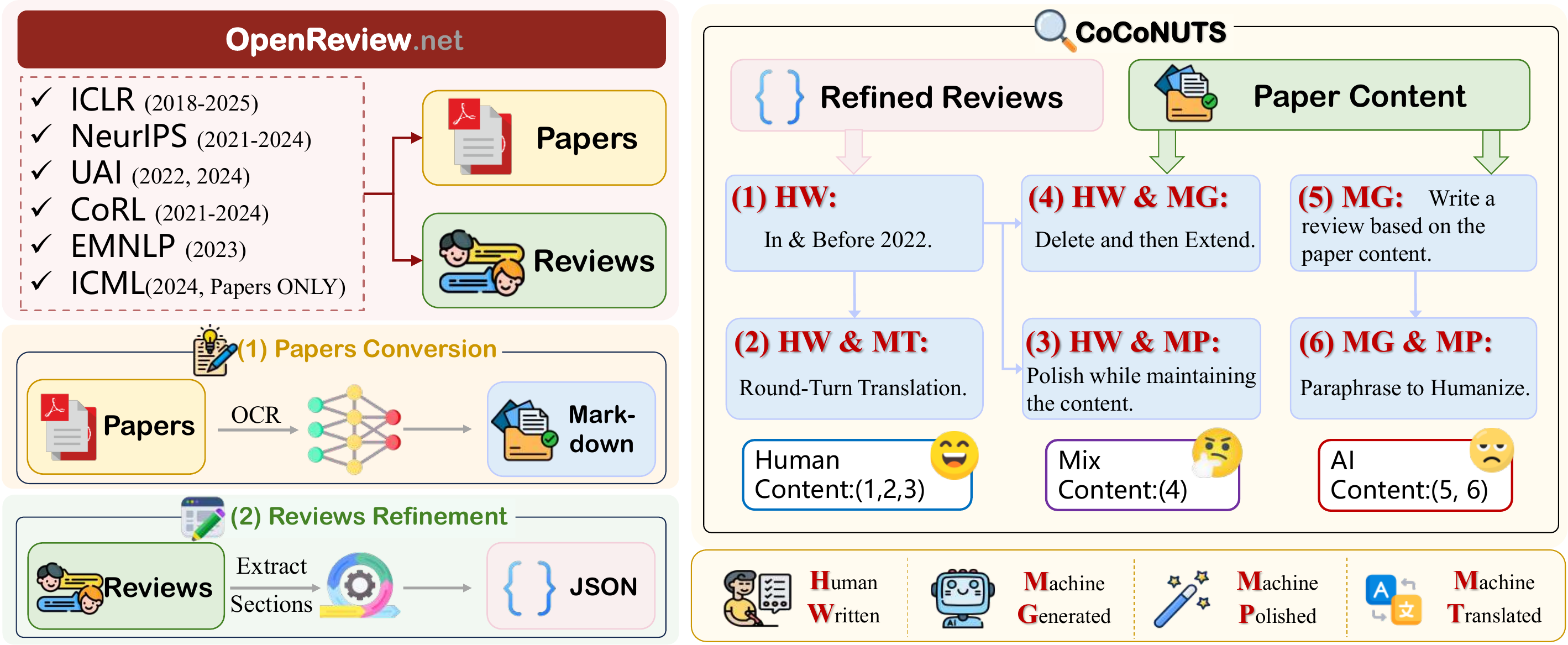}
    \caption{Overview of our CoCoNUTS benchmark. The left side illustrates data acquisition and preprocessing, while the right side shows the construction process for each category and the ternary detection task based on content composition. 
    }
    \label{fig:main}
\end{figure*}

Current general AI-generated text detectors face a dual challenge. They are vulnerable to paraphrasing attacks, allowing humanized AI-generated content to go undetected~\cite{sadasivan2025aigeneratedtextreliablydetected,zhou-etal-2024-humanizing}. Simultaneously, they exhibit high false-positive rates on text with even minor AI polishing, unjustly flagging permissible use~\cite{saha2025almost}. Given that paraphrasing is a semantics-invariant operation~\cite{su2024hc3plussemanticinvarianthuman} that transfers textual style without altering substantive content, these failures suggest a fundamental focus on textual style while neglecting content. Although textual style is a distinguishable feature of AI writing~\cite{NEURIPS2023_575c4500Paraphrasingevades}, this stylistic dependency is particularly problematic in the peer review context, as it risks both unjustly penalizing legitimate AI assistance and overlooking deceptively generated reviews.

To this end, we advocate a paradigm shift in AI-generated review detection by emphasizing content composition over superficial textual style.
First, we introduce CoCoNUTS, a comprehensive peer review benchmark featuring six realistic human-AI collaboration modes, which are in turn categorized into three classes based on their content composition: Human, Mix, and AI.
Second, we propose CoCoDet, a content-concentrated detector. To disentangle content features from stylistic cues, CoCoDet is trained with a multi-task framework, comprising a primary content composition identification task and three  auxiliary tasks.
Together, CoCoNUTS and CoCoDet establish a content-centric framework for reliable AI-generated review detection.

Building on this, we conducted a comprehensive evaluation of a wide range of AI-generated text detectors on the CoCoNUTS benchmark. 
Our results reveal that LLM-based detectors, even with few-shot prompting, struggle to focus on substantive content and tend to rely on superficial stylistic cues, leading to unreliable predictions.
Similarly, general detectors perform poorly in this content-based task, especially non-style-robust models, which completely fail to produce reliable predictions.
CoCoDet achieves state-of-the-art performance, with a macro F1-score exceeding 98\% on the ternary detection task, significantly outperforming both large language models and general detectors. 
Furthermore, when applied to real-world conference reviews, CoCoDet reveals a clear year-over-year increase in AI usage, encompassing not only the now-common practice of AI-assisted polishing but also a growing proportion of fully machine-generated reviews. This trend underscores the practical necessity of adopting robust, content-based detection methods.
Our contributions can be summarized as follows:
\begin{itemize}
\item We introduce a content-centric detection paradigm and present CoCoNUTS, a fine-grained benchmark capturing diverse human-AI collaboration modes.
\item We propose CoCoDet, a robust detector based on a multi-task learning framework that excels at disentangling content from style, achieving state-of-the-art detection performance.
\item We conduct a comprehensive evaluation of existing detectors and reveal the prevalence and different modes of AI involvement in real-world peer review.
\end{itemize}

\section{CoCoNUTS Benchmark} 
To facilitate a fair and robust evaluation of LLM involvement in academic peer review, we introduce CoCoNUTS. The detailed dataset construction and evaluation tasks are illustrated in Figure \ref{fig:main}.

\begin{table*}[t!]
  \centering

  \small

  \begin{tabular}{@{}ll r cccccc cc@{}}
    \toprule
 
    \multicolumn{2}{c}{\multirow{2}[2]{*}{Benchmark}} & \multirow{2}[2]{*}{Size} & \multicolumn{6}{c}{Category} & \multicolumn{2}{c}{Attribution} \\
    
    \cmidrule(lr){4-9} \cmidrule(lr){10-11} 
    \multicolumn{2}{c}{} & & HW & MG & HWMG & HWMT & HWMP & MGMP & Coupled & Separate  \\
    \midrule

    \multicolumn{11}{c}{\textbf{General}} \\
    \midrule
    \multicolumn{2}{l}{MGTBench~\cite{MGTBench}} & 2.82k & \ding{51} & \ding{51} & \ding{55} & \ding{55} & \ding{55} & \ding{51} & \ding{51} & \ding{55}  \\
    \multicolumn{2}{l}{M4~\cite{wang-etal-2024-m4}} & 122k & \ding{51} & \ding{51} & \ding{55} & \ding{55} & \ding{51} & \ding{55} & \ding{51} & \ding{55}  \\
    \multicolumn{2}{l}{HC3 Plus~\cite{su2024hc3plussemanticinvarianthuman}} & 210k & \ding{51} & \ding{55} & \ding{55} & \ding{51} & \ding{51} & \ding{55} & \ding{55} & \ding{55}  \\
    \multicolumn{2}{l}{RAID~\cite{dugan-etal-2024-raid}} & 6.2M & \ding{51} & \ding{51} & \ding{55} & \ding{55} & \ding{55} & \ding{51} & \ding{51} & \ding{55}  \\
    \multicolumn{2}{l}{MixSet~\cite{zhang-etal-2024-llmascoauthor}} & 3.6k & \ding{51} & \ding{51} & \ding{51} & \ding{55} & \ding{51} & \ding{51} & \ding{51} & \ding{55}  \\
     \multicolumn{2}{l}{M4GT-Bench~\cite{wang-etal-2024-m4gt}} & 217k & \ding{51} & \ding{51} & \ding{51} & \ding{55} & \ding{55} & \ding{55} & \ding{51} & \ding{55}  \\
    \multicolumn{2}{l}{CUDRT~\cite{tao2024cudrt}} & 480k & \ding{51} & \ding{51} & \ding{51} & \ding{51} & \ding{51} & \ding{55} & \ding{51} & \ding{55}  \\
    \multicolumn{2}{l}{LAMP~\cite{chakrabarty2025canLAMP}} & 1.06k & \ding{51} & \ding{51} & \ding{51} & \ding{55} & \ding{55} & \ding{51} & \ding{51} & \ding{55}  \\
    \multicolumn{2}{l}{Beemo~\cite{artemova-etal-2025-beemo}} & 19.6k & \ding{51} & \ding{51} & \ding{51} & \ding{55} & \ding{55} & \ding{51} & \ding{51} & \ding{55}  \\
   
    \multicolumn{2}{l}{HART~\cite{bao2025decouplingcontentexpressiontwodimensionalhart}} & 32k & \ding{51} & \ding{51} & \ding{55} & \ding{55} & \ding{51} & \ding{51} & \ding{55} & \ding{55}  \\
    \multicolumn{2}{l}{OpenTuringBench~\cite{lacava2025openturingbenchopenmodelbasedbenchmarkframework}} & 543k & \ding{51} & \ding{51} & \ding{51} & \ding{55} & \ding{51} & \ding{51} & \ding{51} & \ding{55}  \\
    \midrule

    \multicolumn{11}{c}{\textbf{News}} \\
    \midrule
    \multicolumn{2}{l}{TuringBench~\cite{uchendu-etal-2021-turingbench-benchmark}} & 200k & \ding{51} & \ding{51} & \ding{55} & \ding{55} & \ding{55} & \ding{55} & \ding{51} & \ding{55}  \\
    \multicolumn{2}{l}{LLMDetect~\cite{cheng2025beyondbinary}} & 64.3k & \ding{51} & \ding{51} & \ding{51} & \ding{55} & \ding{51} & \ding{55} & \ding{51} & \ding{55}  \\
    \midrule


    \multicolumn{11}{c}{\textbf{Academic}} \\
    \midrule
    \multicolumn{2}{l}{FAIDset~\cite{ta2025faidfinegrainedaigeneratedtext}} & 83.3k  & \ding{51} & \ding{51} & \ding{51} & \ding{55} & \ding{51} & \ding{51} & \ding{51} & \ding{55}  \\
    \multicolumn{2}{l}{AIPR-Detection-Benchmark~\cite{yu2025isyourpaper}} & 789k & \ding{51} & \ding{51} & \ding{55} & \ding{55} & \ding{55} & \ding{55} & \ding{51} & \ding{55}  \\
    \multicolumn{2}{l}{\textbf{CoCoNUTS (Ours)}} & 316k & \ding{51} & \ding{51} & \ding{51} & \ding{51} & \ding{51} & \ding{51} & \ding{51} & \ding{51}  \\

    \bottomrule
  \end{tabular}%

    \caption{Comparison of publicly available resources for AI-Generated Text detection. Our work is the only resource that includes texts from six distinct usage scenarios. Furthermore, we are the first to propose disentangled model attribution, with separate labels for the initial content generator and the final style converter.}
    \label{tab:dataset comparison}
\end{table*}%

\subsection{Dataset Construction}
To address the limitations of existing datasets in representing diverse AI use in peer review, we constructed a large-scale dataset of 315,535 instances, comprising six  categories designed to simulate realistic human-AI collaboration modes. The detailed construction process is as follows.

First, we collected reviews and their corresponding papers from OpenReview, covering venues including ICLR (2018–2025), NeurIPS (2021–2024), UAI (2022, 2024), CoRL (2021–2024), and EMNLP (2023). To enhance the diversity of generated reviews, we incorporated papers from ICML 2024, whose reviews were not publicly available.
From the reviews, we extracted only the substantive sections, such as the main analysis and specific questions for the authors, while discarding templated content like rating and confidence. 
This step purifies the data by focusing on substantive content and eliminating variations from different review forms. Concurrently, all collected papers were converted from PDF to Markdown format using Nougat~\cite{blecher2024nougat} to support further processing.

Next, we constructed the six data categories using a carefully designed generation pipeline. This pipeline employs a suite of advanced LLMs, including DeepSeek-R1-671B, Gemini-2.5-flash-0520, Llama-3.3-70B-Instruct, Qwen2.5-72B-Instruct, and Qwen3-32B. The construction methods for each category are as follows:
\begin{description}
    \item[HW (Human-Written)] To ensure the purity of the data, we selected reviews written in or before 2022 from our collection, which is prior to the public release of ChatGPT. 
    \item[HWMT (Human-Written \& Machine-Translated)] We applied back-translation to HW reviews using LLMs, translating them into Chinese and then back to English to introduce stylistic variation.
    \item[HWMP (Human-Written \& Machine-Polished)] We polished the HW reviews using LLMs without altering the core meaning.
    \item[HWMG (Human-Written \& Machine-Generated)] We provided LLMs with both an original HW review and its corresponding paper sections. The model was prompted to prune redundant parts from the original review and add critical points that were missed.
    \item[MG (Machine-Generated)] We provided the LLMs with several HW reviews, followed by the content of a source paper (variably the full text or key sections), and tasked it with generating a review.
    \item[MGMP (Machine-Generated \& Machine-Polished)] We processed the MG reviews with a different LLM prompted to paraphrase them to simulate a user trying to humanize the content to evade AI detection.
\end{description}

Finally, to ensure data quality, we first eliminated conversational phrases (e.g., ``Here is the polished review''). We then removed  samples that fell outside the 5th-95th percentile token length of the HW set to address uninformatively short or excessively long reviews. The effectiveness of the process was validated by manually inspecting a random sample of 100 instances.  For the machine-generated (MG) category, we further incorporated reviews generated by Claude 3.5 Sonnet and GPT-4o from the AIPR-Detection-Benchmark~\cite{yu2025isyourpaper}. The final dataset statistics are presented in Table~\ref{tab:FAIRset}.

\begin{table}[h!]
  \centering
  \small
  {
    \setlength{\tabcolsep}{1pt}
    \small
    \begin{tabular}{@{}cccc@{}}
      \toprule
      \multicolumn{4}{l}{\textbf{Human} \hfill 105,180} \\
      \cmidrule(lr){1-1} \cmidrule(lr){2-3}
      \multicolumn{1}{c}{\textbf{HW}} & \multicolumn{2}{c}{\textbf{HWMT}} & \\
      Human & Llama & Qwen2.5 &  \\
      35,060 & 17,142 & 17,918 &  \\
      \cmidrule(lr){1-4}
      \multicolumn{4}{c}{\textbf{HWMP}} \\
      Gemini & Llama & Qwen3 & \\
      10,372 & 12,450 & 12,238 &  \\
      \midrule 
      
      \multicolumn{4}{l}{\textbf{Mix} \hfill 105,180} \\
      \cmidrule(lr){1-4} 
      \multicolumn{4}{c}{\textbf{HWMG}} \\
      Gemini & Llama & Qwen2.5 & Qwen3 \\
      19,251 & 43,997 & 29,201 & 12,731 \\
      \midrule 

      \multicolumn{4}{l}{\textbf{AI} \hfill 105,175} \\
      \cmidrule(lr){1-4} 
      \multicolumn{4}{c}{\textbf{MG}} \\
      Claude* & DeepSeek & Gemini & GPT* \\
      7,578  & 3,500  & 8,000 & 7,597 \\
      Llama & Qwen2.5 & Qwen3 &  \\
      13,500 & 10,000 & 10,000 &  \\
      \cmidrule(lr){1-4}
      \multicolumn{4}{c}{\textbf{MGMP}} \\
      Gemini-Llama & Llama-Gemini & Llama-Qwen2.5 & DS-Gemini  \\
      7,000 & 8,000 & 7,000 & 3,000 \\
      DS-Llama & Qwen2.5-Gemini & Qwen3-Gemini & Qwen3-Llama \\
      5,000 & 4,000 & 5,000  & 6,000 \\
      \bottomrule
    \end{tabular}%
  } 
    \caption{Dataset statistics. * Data sourced from the AI-Peer-Review-Detection-Benchmark.}
      \label{tab:FAIRset}
\end{table}%
\subsection{Task Definition}
The task of AI text detection faces a fundamental trade-off between expressiveness and robustness. Binary classification~\cite{guo2023howclosechatgpthumanexperts} lacks the expressiveness for the scenarios of human-AI co-authorship, while fine-grained paradigms, such as model-specific attribution~\cite{ta2025faidfinegrainedaigeneratedtext} are brittle against unseen models and collaboration patterns. To resolve this dilemma, we introduce a ternary classification based on the content composition. We posit that semantic-invariant operations alter textual style without changing substantive content (see Appendix~\ref{appendix:semanticinv} for analysis). Based on this principle, we map our six granular categories into three content-based classes. Specifically, reviews with purely human content, even after such operations (HW, HWMT, HWMP), are labeled as ``Human''. Conversely, reviews with purely AI content, including paraphrased versions (MG, MGMP), are labeled as ``AI''. The co-authored scenario (HWMG) with content from both human and AI, is designated as ``Mix''. This ternary setup serves as the basis for our main detection task and evaluation.

\section{CoCoDet Detector} 


General detectors are prone to misclassify text paraphrased in human-AI collaboration due to their reliance on style. To address this, we introduce the Content-Concentrated Detector (CoCoDet) with a tailored multi-task training framework. 
To enable the model to learn robust  representations, this framework integrates a primary task, Content Composition Identification from the CoCoNUTS benchmark, with three carefully designed auxiliary tasks: Collaboration Mode Attribution, Content Source Attribution, and Textual Style Attribution. These auxiliary tasks enable the model to separate content features from stylistic ones, allowing for a reliable classification of the content composition.

\paragraph{Content Composition Identification:}
The primary task, predefined by CoCoNUTS benchmark, requires identifying a review's substantive origin into three classes based on its content composition: \textit{Human}, \textit{AI}, or \textit{Mix}. 
This content-concentrated approach aims to guide the model towards a fair and robust detection.
A key challenge in this task is ensuring clear decision boundaries. 
To this end, we first adopt the large margin cosine loss to enhance class separability~\cite{Wang_2018_cosface}. 
Let $\mathbf{x}$ be the feature embedding and $\mathbf{r}_j$ be the  weight vector for class $j$. The logit for class $j$ is the cosine similarity $z_j = \cos(\theta_j)$, where $\theta_j$ is the angle between $\mathbf{x}$ and $\mathbf{r}_j$. A base margin $m_{\text{base}}$ is subtracted from the logit of the ground-truth class $y$:
\begin{equation}
\label{eq:base_margin_logit}
z'_j = 
\begin{cases} 
z_j - m_{\text{base}} & \text{if } j = y \\
z_j & \text{if } j \neq y
\end{cases}
\end{equation}

Furthermore, considering that confusing \textit{Human} and \textit{AI} texts incurs a significantly higher cost in our context, we introduce an additional cost margin, $m_{\text{cost}}$, which modifies the logits of these critical negative classes during training. We formalize this entire mechanism as the \textbf{Cost-Sensitive Margin Loss (CSM-Loss)}. Based on the intermediate logits $z'_j$, we apply the cost margin $m_{\text{cost}}$ to produce the final logits:
\begin{equation}
z_j^* \!=\!
\begin{cases} 
s \!\cdot\! (z'_j \!+\! m_{\text{cost}}) & \text{if } j,y \!\in\! \{\text{human, ai}\} \text{ and } j \!\neq\! y \\
s \!\cdot z'_j & \text{otherwise}
\end{cases}
\end{equation}
where $s$ is a scaling parameter. 
This targeted penalty structure compels the model to learn a more discriminative feature representation. Specifically, our CSM-Loss encourages 
inter-class separability, especially between the high-cost Human and AI classes. 
For example, correct classification decision for a human-class review requires that:
\begin{equation}
z_{\text{human}} > \max(z_{\text{ai}} + m_{\text{base}} + m_{\text{cost}},z_{\text{mix}} + m_{\text{base}})
\end{equation}

Let $C$ be the three classes. The final loss $\mathcal{L}_{main}$ is the cross-entropy loss over these modified logits $z^*_j$:
\begin{equation}
\label{eq:final_loss_main}
\mathcal{L}_{\text{main}} = -\log\left(\frac{e^{z_y^*}}{\sum_{k=1}^{C} e^{z_k^*}}\right)
\end{equation}

\paragraph{Content Source Attribution:}
This multi-label classification task aims to trace the substantive content back to its true origin by identifying the specific model that performed the initial generation. We operate on the hypothesis that human experts and different AI models possess unique capabilities and preferences (e.g., knowledge cutoffs, areas of focus). These intrinsic differences lead to discernible, author-specific characteristics in the content they generate. By training the model to attribute content to its initial author, we compel it to move beyond superficial stylistic analysis and instead learn to identify who is capable of producing what kind of substantive critique. To capture these specific traits, we employ a direct, one-to-one mapping for the labels (e.g.,``Qwen2.5'' and ``Qwen3'' are treated as distinct labels). For a review first generated by Qwen3 and then polished by Gemini, the ground-truth label is ``Qwen3''. And for a Mix review that was edited by Qwen3, the labels would be [``Human'', ``Qwen3''] to reflect the dual contribution. 
The final loss $\mathcal{L}_{\text{con}}$ is the binary cross-entropy (BCE) with logits: 
\begin{equation}
\label{eq:loss_content}
\mathcal{L}_{\text{con}}\! = \!-\frac{1}{n}\! \sum_{i=1}^{n}  \left[ y_i \log(\sigma(u_i))\!+\! (1\!-\!y_i) \log(1\!-\!\sigma(u_i)) \right]
\end{equation}
where $n$ is the number of content source labels, $y_i$ is the binary ground-truth, and $u_i$ is the output logits from the final linear layer for content source attribution.

\paragraph{Textual Style Attribution:}
This task seeks to identify the model responsible for the textual style of a review in order to enable the model to capture the stylistic patterns indicative of the final authoring or editing model. In conjunction with the Content Source Attribution, this task enables the model to disentangle what is said (content) from how it is said (style). 
In this task, based on the rationale that models from the same family develop consistent stylistic features, we group models into families (e.g., Qwen2.5 and Qwen3 are mapped to the ``Qwen'' label). For a review  generated by Llama then polished by Qwen2.5, the ground-truth label would be ``Qwen''. This loss $\mathcal{L}_{\text{sty}}$ is also the BCE with logits:
\begin{equation}
\label{eq:loss_style}
\mathcal{L}_{\text{sty}}\! = \!-\frac{1}{m}\! \sum_{i=1}^{m}  \left[ y_i \log(\sigma(v_i))\!+\! (1\!-\!y_i) \log(1\!-\!\sigma(v_i)) \right]
\end{equation}
where $m$ is the number of style labels, $y_i$ is the binary ground-truth, and $v_i$ is the output logits from the final linear layer for textual style attribution.

\paragraph{Collaboration Mode Attribution:}
This multi-class classification task compels the model to understand the fine-grained compositional provenance of a text by attributing it to a specific collaboration mode. By classifying each review into one of the six predefined modes in the CoCoNUTS dataset (e.g., HW, HWMP, and MGMP), the model is explicitly informed about the latent hierarchy of content composition. The loss $\mathcal{L}_{\text{mode}}$ is the cross-entropy loss over logits:
\begin{equation}
\label{eq:loss_mode}
\mathcal{L}_{\text{mode}} = -\log\left(\frac{e^{w_y}}{\sum_{j=1}^{M} e^{w_j}}\right)
\end{equation}
where $M$ is the total number of collaboration modes, $w_y$ is the output logit from the final linear layer for the ground-truth class $y$, and $w_j$ is the logit for the $j$-th class.

Based on these defined training tasks, we adopt ModernBERT~\cite{warner2024modernbert} as the backbone of our CoCoDet detector. The model is trained end-to-end using a composite loss function that linearly combines the loss from the primary task with weighted losses from the three auxiliary tasks.
The composite loss $\mathcal{L}$ is defined as:
\begin{equation}
\mathcal{L} = \mathcal{L}_{main} + \alpha\mathcal{L}_{con} + \beta\mathcal{L}_{sty} + \gamma\mathcal{L}_{mode}
\end{equation}
where the scalar weights $\alpha$, $\beta$, and $\gamma$ are hyperparameters.

\section{Experiments} 

We conducted a series of experiments on the CoCoNUTS benchmark to evaluate the CoCoDet detector. 
First, we compared CoCoDet with LLM-based detectors to show its effectiveness. To further reveal the limitations of general detectors, we evaluated their binary classification performance across the Human, Mix, and AI subsets of CoCoNUTS. 
We then performed an ablation study to validate the components of CoCoDet. 
Finally, we applied CoCoDet to analyze AI usage trends in real-world post-ChatGPT peer reviews.

\subsection{Experimental Setup}
\paragraph{Dataset:}
We partition the dataset into training, validation, and test sets with an 8:1:1 ratio using stratified random sampling. This ensures that the class distribution is consistent across all splits (see Appendix~\ref{appendix:dataspilt} for details). All final performance metrics are reported on the held-out test set.

\paragraph{Baselines:}
We benchmark CoCoDet against nine baseline detection methods, including four recent advanced LLMs  and five mainstream general AI-generated text detectors.

For the \textit{LLM-based detectors}, we selected DeepSeek-R1-0528, Gemini-2.5-flash-0520, Qwen2.5-72B-Instruct, and Qwen3-32B as baseline models. 
Their performance was evaluated in both zero-shot and few-shot settings. Notably, for Gemini-2.5-Flash, we evaluated its performance in both its thinking and non-thinking modes. We prompted the models to determine the review's origin by focusing on its substantive content composition rather than its stylistic features(see Appendix~\ref{appendix:prompt4llmdetector} for detailed prompts), and choose one from the three options: \textit{Human}, \textit{AI}, or \textit{Mix}. 
In the few-shot scenario, we provided each model with one in-context example per class, ensuring the set of examples was identical for all models for a fair comparison. 

For the \textit{general detectors}, we selected five mainstream methods, encompassing both model-based and metric-based approaches. 
The model-based methods include Radar~\cite{hu2023radar} and LLM-DetectAIve~\cite{abassy-etal-2024-llmdetectaive}. 
The metric-based methods include LLMDet~\cite{wu2023llmdet}, FastDetectGPT~\cite{bao2024fastdetectgpt} and Binoculars~\cite{hans2024spotting}. We strictly adhered to the officially recommended configurations for all general detectors. Model-based methods were run on the same hardware, while metric-based methods utilized their prescribed thresholds comparison. For Binoculars, we report results at both its accuracy and low-false-positive-rate (low-fpr) thresholds.

\paragraph{Training Configuration:}

We fine-tuned CoCoDet with the AdamW optimizer~\cite{loshchilov2018adamw}, selecting the best model on the validation set. Hyperparameters were optimized via a sequential grid search (detailed in Appendix~\ref{appendix:traininghyper}), where we successively tuned: 
(1) the learning rate over \{1e-5, 2e-5, \dots, 5e-5\}; 
(2) the base and cost margins $m_{\text{base}}, m_{\text{cost}}$ from \{0.2, 0.25, 0.3\}; 
and (3) the auxiliary task weights, by first jointly searching $\alpha$ and $\beta$ over \{0.2, 0.3, 0.4, 0.5\} and then tuning $\gamma$ over \{0.1, 0.2, 0.3, 0.4\}. 
All experiments used a fixed random seed of 42 for reproducibility.



\paragraph{Evaluation Metrics:}
For LLM-based detectors, we report per-class F1-scores for the Human, Mix, and AI classes, along with the average F1-score. 
For general detectors, we report the predicted AI rate on each class and the average accuracy, defined as the mean accuracy on the Human and AI classes. 
To facilitate a fair comparison, we mapped the outputs of multi-class models. For the four-way classifier LLM-DetectAIve, we mapped the ``HW/HWMP'' predictions  to ``Human'' and ``MG/MGMP'' predictions to ``AI''. For CoCoDet, we applied a more stringent standard: any non-Human prediction on human subset  was considered a false positive, while  only a  AI prediction was counted as a true positive on AI subset.

\begin{table}[t] 
\centering
\small
\setlength{\tabcolsep}{2pt} 
\begin{tabular}{
    @{}
    l
    S[table-format=2.2]
    S[table-format=2.2]
    S[table-format=2.2]
    S[table-format=2.2]
    @{}
} 
\toprule
\textbf{Detector} & 
{\textbf{Human}} & 
{\textbf{Mix}} & 
{\textbf{AI}} & 
{\textbf{Average}} \\
\midrule

\multicolumn{5}{@{}l}{\textbf{LLMs (zero-shot)}} \\
\quad DeepSeek-R1-0528            & 50.04 & 3.29 & 3.63 & 18.98 \\
\quad Gemini-2.5-flash-0520(CoT)  & 56.01 & 2.81 & 47.87 & 35.56 \\
\quad Gemini-2.5-flash-0520       & 57.28 & 12.37 & 49.80 & 39.82 \\
\quad Qwen2.5-72B-Instruct    & 48.47 & 3.05 & 16.82 & 22.78 \\
\quad Qwen3-32B                   & 50.30 & 0.11 & 4.89 & 18.43 \\
\midrule

\multicolumn{5}{@{}l}{\textbf{LLMs (few-shot)}} \\
\quad DeepSeek-R1-0528            & 51.81 & 5.65  & 17.93 & 25.13 \\
\quad Gemini-2.5-flash-0520 (CoT)  & 64.95 & 10.87 & 61.42 & 45.75 \\
\quad Gemini-2.5-flash-0520       & 74.05 & 39.90 & 62.97 & 58.97 \\
\quad Qwen2.5-72B-Instruct    & 47.17 & 16.85 & 14.61 & 26.21 \\
\quad Qwen3-32B                   & 53.64 & 0.02  & 38.39 & 30.68 \\
\midrule
\multicolumn{5}{@{}l}{\textbf{PLM (SFT)}} \\
\quad \textbf{CoCoDet}                      & \textbf{98.94} & \textbf{97.41} & \textbf{98.37} & \textbf{98.24} \\ 
\bottomrule
\end{tabular}
\caption{Performance on the ternary classification task. We report the per-class and average F1-scores (\%).}
\label{tab:ternary_f1_results}
\end{table}

\subsection{Overall Results}

\paragraph{Large language models are ill-suited for the task of content-concentrated detection.}
As detailed in Table~\ref{tab:ternary_f1_results}, LLM-based detectors struggle to achieve reliable results on the CoCoNUTS benchmark. In a zero-shot setting, their performance is notably poor. Despite an improvement from few-shot prompting over a poor zero-shot baseline, the overall efficacy of LLM-based detectors remains severely limited. Even the best LLM fails to surpass a 60\% average F1-score, while the majority of other models operate near or below chance levels. An unexpected finding is the degraded performance of Gemini in its thinking mode compared to its non-thinking mode. To investigate the cause, we analyzed the reasoning processes of Qwen3 and DeepSeek, as Gemini's reasoning process is inaccessible. We found that despite being explicitly prompted to judge based on substantive content, the reasoning frequently defaulted to analyzing textual style(e.g., overly polished transitions or formulaic phrasing). This tendency to fixate on stylistic cues, even when instructed otherwise, appears to be a key factor limiting the performance of LLMs on this task.

\begin{table}[t]
\centering
\small
\setlength{\tabcolsep}{2pt} 
\begin{tabular}{@{} l S[table-format=2.2] S[table-format=2.2] S[table-format=2.2] S[table-format=2.2] S[table-format=2.2] @{}}
\toprule
\multirow{2}{*}{\textbf{Detector}} & \multicolumn{3}{c}{\textbf{Predicted AI Rate }} & {\multirow{2}{*}{\textbf{Acc$\uparrow$}}} & {\multirow{2}{*}{\textbf{Sty-Rob}}}  \\
\cmidrule(l){2-4}
& {\textbf{Human}$\downarrow$} & {\textbf{Mix}} & {\textbf{AI}$\uparrow$}  \\
\midrule
Radar & 24.91 & 26.33 & 34.93 & 55.01 &\ding{51} \\
LLMDet & 98.82 & 98.45 & 99.26 & 50.22 &\ding{55} \\
FastDetectGPT & 53.09 & 92.98 & 92.56 & 69.74 &\ding{55} \\
Binoculars(accuracy) & 15.86 & 66.96 & 74.32 & 79.23 &\ding{51} \\
Binoculars(low-fpr) & 3.30 & 34.78 & 49.81 & 73.26 &\ding{51} \\
LLM-DetectAIve & 3.92 & 33.89 & 83.52 & \bfseries 89.80 &\ding{51} \\
\textbf{CoCoDet} & {\bfseries \enspace 1.31} & {--} & {\bfseries 96.90} & {\bfseries 97.80} &{--}\\
\bottomrule
\end{tabular}
\caption{Performance of general detectors . The Acc shows the mean accuracy for Human and AI sets (\%). The Sty-Rob indicates the style robustness of the binary detector.}
\label{tab:binary_detection_performance}
\end{table}

\begin{figure}[t]
    \centering
    \includegraphics[width=0.95\linewidth]{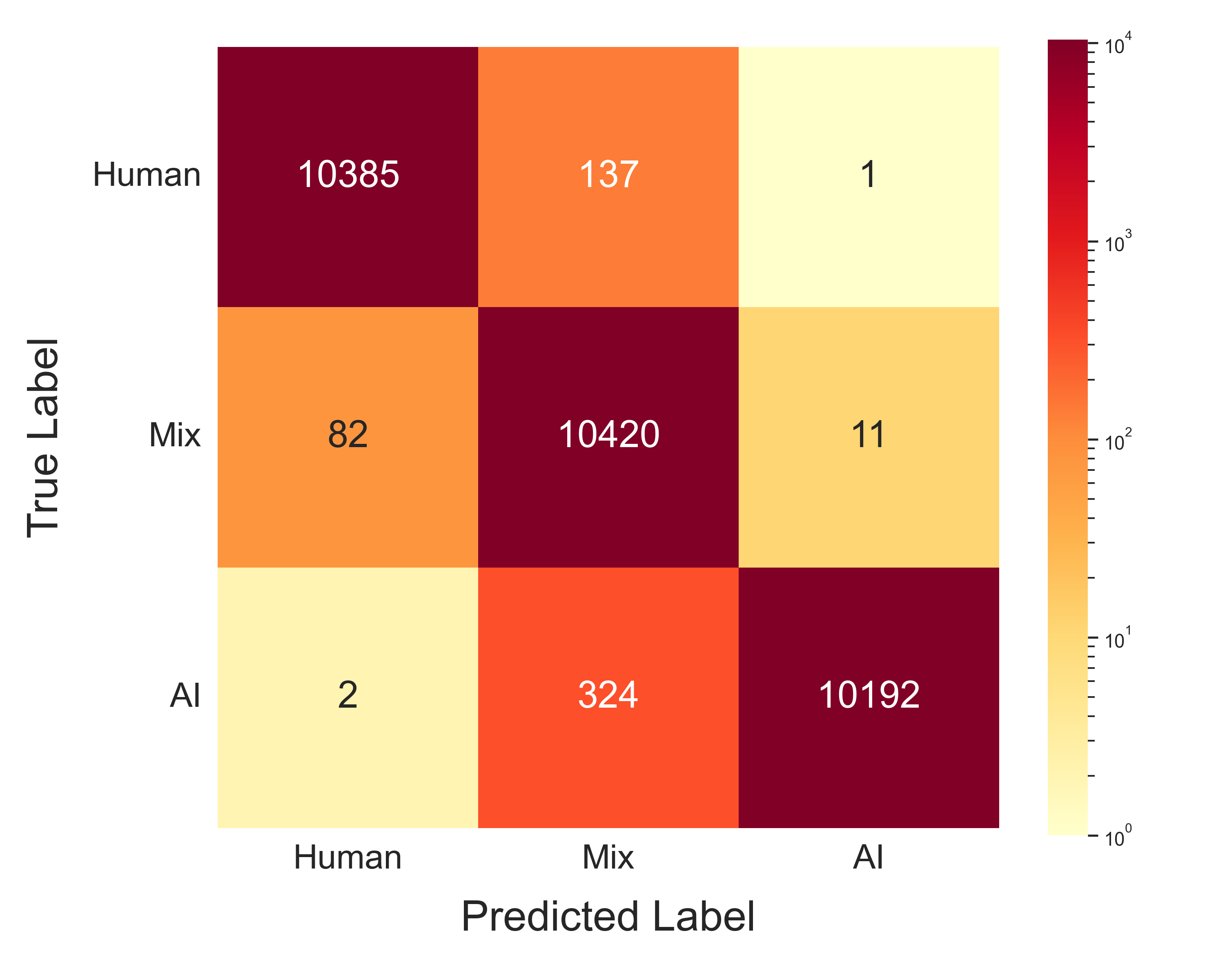}
    \caption{Confusion matrix of CoCoDet on the ternary classification task. The model exhibits high accuracy, with predictions clustered along the main diagonal. Misclassifications are mostly confined to adjacent classes, while critical errors between the distinct Human and AI classes are rare.}
    \label{fig:confusion}
\end{figure}

\paragraph{General detectors' performance is dictated by style-robustness.}
Our benchmark against general detectors, presented in Table~\ref{tab:binary_detection_performance}, reveals significant performance disparities and an overall inadequacy on this task. None of the general detectors achieve an average accuracy of 90\%, and their classification of mix samples lacks uniformity. To explain these results, we introduce style-robustness metric, defining a detector as style-robust if its Predicted AI Rate monotonically increases from the Human, to the Mix, and finally to the AI subset. This metric reflects the ability to classify based on substantive content composition rather than superficial stylistic features. 
Methods explicitly designed to be style-robust, such as Radar, Binoculars, and LLM-DetectAIve successfully meet our monotonicity criterion.
We observe a correlation between this style-robustness and accuracy. The robust detectors, LLM-DetectAIve and Binoculars, outperform those non-robust like LLMDet and FastDetectGPT. A notable exception is Radar. While style-robust, it shows limited accuracy, which we attribute to its lack of training data from the review domain.

\paragraph{CoCoDet achieves a comprehensive state-of-the-art detection performance.}
CoCoDet achieves state-of-the-art results across all evaluated metrics, demonstrating its superiority over different classes of baseline models. Compared with LLM-based detectors, CoCoDet achieves a macro F1-score of 98.24\% as seen in Table~\ref{tab:ternary_f1_results}, drastically outperforming the best-performing few-shot LLM by nearly 40 percentage points. When benchmarked against general detectors, it also secures a top accuracy of 97.80\% as seen in Table~\ref{tab:binary_detection_performance}, coupled with an exceptionally low false positive rate on human text of a mere 1.31\%. Further analysis of the confusion matrix (Figure~\ref{fig:confusion}) reveals that critical errors between the \textit{Human} and \textit{AI} classes are virtually eliminated. It directly validates the effectiveness of our CSM Loss, which is specifically designed to penalize these severe errors.

\subsection{Ablation Study}
To validate the efficacy and contribution of each component of CoCoDet, we conduct a comprehensive ablation study. We systematically create several ablated versions of our model by removing key components one at a time: the auxiliary task losses and the core elements of our main loss function.
To ensure a fair comparison, the experimental setup for each ablated model is kept strictly identical to that of the full CoCoDet model. This includes fine-tuning for 5 epochs, with all other hyperparameters remaining unchanged. For each variant, we select the model checkpoint that achieves the best performance on the validation set for the final evaluation. The ablation results are detailed in Table \ref{tab:ablation_study}.

\begin{table}[t]
  \centering
  \small
  \setlength{\tabcolsep}{3pt} 
  \begin{tabular}{lrrrr} 
    \toprule
    \textbf{Model Configuration} & 
    \multicolumn{1}{c}{\textbf{Human}} & 
    \multicolumn{1}{c}{\textbf{Mix}} & 
    \multicolumn{1}{c}{\textbf{AI}} & 
    \multicolumn{1}{c}{\textbf{Average}} \\
    \midrule
    CoCoDet (Full Model) & 98.94 & 97.41 & 98.37 & 98.23 \\
    \midrule
    \multicolumn{4}{l}{\textit{Ablation of Auxiliary Tasks:}} \\
    \quad -- Content Source   & 98.73 & 96.74 & 97.77 & 97.75 \\
    \quad -- Textual Style   & 98.72 & 96.74 & 97.86 & 97.77 \\ 
    \quad -- Collaboration Mode   & 98.73 & 95.77 & 96.44 & 96.98 \\
    \midrule
    \multicolumn{4}{l}{\textit{Ablation of Main Task:}} \\
    \quad -- Base Margin & 99.17 & 95.02 & 95.40 & 96.53 \\
    \quad -- Cost Margin & 98.87 & 94.98 & 95.55 & 96.47 \\
    \bottomrule
  \end{tabular}
    \caption{Ablation study of CoCoDet demonstrates the benefit of each component of the framework in F1-score (\%)}
  \label{tab:ablation_study}
\end{table}

\textbf{All auxiliary tasks contribute positively to the final performance.} 
The observation that removing either the content source attribution or textual style attribution individually results in a modest performance drop is not a sign of redundancy. Instead, it highlights their complementary nature in creating an implicit disentanglement, where learning to isolate content inherently helps to identify style, and vice versa. While the model exhibits the capability to infer one representation from the other, the full model's superior performance demonstrates that explicit dual supervision is crucial. In contrast, the more substantial decline observed upon removing the collaboration mode attribution validates its distinct and orthogonal role. 

\textbf{Both margin factors in main task contributes to the detection effectiveness.}
Removing the Base Margin significantly impairs performance, indicating that a universal large decision boundary is fundamental for preventing classification ambiguity.
The Cost Margin proves even more vital, as its removal results in the most significant performance degradation across all experiments. The targeted penalty of the cost margin is essential for resolving the critical distinction between purely human and AI classes. 


\subsection{AI Usage Trends in Post-ChatGPT Reviews}
\begin{figure}[t]
    \centering
    \includegraphics[width=\linewidth]{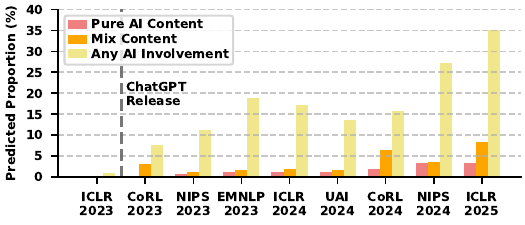}
    \caption{The predicted AI involvement of recent conference reviews. Pure AI Content and Mix Content correspond to the outputs of content composition identification. Any AI Involvement aggregates all categoriesexcept ``HW'' by the collaboration mode attribution.}
    \label{fig:predicted}
\end{figure}

To investigate real-world AI usage trends, we applied CoCoDet to peer reviews from top-tier conferences since 2023, with the results depicted in Figure~\ref{fig:predicted}. As a baseline, we first analyzed reviews from ICLR 2023 (pre-ChatGPT), where CoCoDet yielded a false positive rate below 1\% for Any AI Involvement and zero false positives for the Mix and Pure AI content classes, confirming the reliability.
In contrast, the analysis of post-ChatGPT reviews reveals an escalating trend of AI involvement. The output indicates that the use of AI in peer review has become a widespread phenomenon, which is evidenced by the high proportion of Any AI Involvement across all recent conferences. This involvement primarily consists of AI assistance for language enhancement, as suggested by the significant gap between the Any AI Involvement  and  the other classes. However, a more concerning pattern also emerges: the proportion of Pure AI Content shows a  year-over-year increase, confirming that the irresponsible practice of submitting entirely machine-generated reviews is a real issue. Beyond the visible trends in the data, our manual analysis reveals another common pattern of misuse: the paper summary section often exhibits AI-generated features, whose risks warrant equal attention.

\section{Related work}

\paragraph{AI's Role and Risks in Peer Review:}
Recent research indicates a growing use of LLMs in peer review, where their application extends beyond language polishing to the substantive modification of content~\cite{zhou2025largelanguagemodelspenetration,liang2024monitoring}, noting a correlation between LLM use and reviews submitted near deadlines, suggesting LLMs serve as a compensatory tool under pressure. However, this utility is accompanied by significant risks to academic integrity. Studies consistently show that LLM-generated reviews are often superficial and lack actionable suggestions \cite{zhou-etal-2024-llm, du-etal-2024-llmsassistnlp}.  This degradation in review quality undermines the core function of peer review.

\paragraph{Benchmarks of AI-Generated Text Detection:}
Benchmarks of the detection of AI-generated text have progressed from simple binary classification~\cite{guo2023howclosechatgpthumanexperts} to more nuanced paradigms like source attribution and authorship boundary detection~\cite{dugan-etal-2023-roft}.  Even so, these paradigms still struggle with the deeply intertwined nature of modern human-AI collaborative writing~\cite{zhang-etal-2024-llmascoauthor}. In response, recent efforts have produced fine-grained datasets classifying the role of AI \cite{wang-etal-2024-m4gt}. However, these datasets largely overlook the unique context of academic peer review. A recent work targets this domain and highlights concerns about AI editing~\cite{yu2025isyourpaper}, while its released dataset is confined to a binary scenario.

\paragraph{AI-Generated Text Detectors:}
General AI-generated text detectors are broadly divided into metric-based and model-based approaches. Model-based methods frame detection as a supervised classification task by fine-tuning pretrained language models to learn distinguishing patterns~\cite{abassy-etal-2024-llmdetectaive}. Metric-based methods analyze properties such as conditional probability curvature~\cite{bao2024fastdetectgpt} and perplexity~\cite{hans2024spotting} to differentiate machine output from human writing. Recent work has begun to adapt these methods for peer review detection. For binary classification, researchers proposed methods using term frequency and review regeneration~\cite{kumar2024quiscustodietipsoscustodes} as well as the Anchor embeddin approach~\cite{yu2025isyourpaper}.
For mix reviews, MixRevDetect\cite{kumar-etal-2025-mixrevdetect} introduced a sentence-level detection via LLM completion.
However, these  approaches either rely on fragile stylistic cues or require LLM assistance at inference, incurring extra overhead.

\section{Conclusion} 
In this work, we proposed a paradigm shift that concentrates on the content rather than textual style for AI-generated review detection. To this end, we introduced the CoCoNUTS benchmark and CoCoDet detector trained with a tailored multi-task framework. Our experiments validate the necessity of the paradigm. We found that baseline detectors are ill-suited for the detection of reviews. In contrast, CoCoDet demonstrates robust detection capabilities, achieving outstanding performance with over 98\% macro F1-score. Applying CoCoDet to recent reviews, we revealed an accelerating trend of AI adoption, encompassing not only widespread use for language enhancement but also a concerning rise in fully machine-generated content.
The goal of this work is to provide a basis for attributing the source of AI use within the peer review process and guide the responsible and transparent integration of 
AI into the scholarly ecosystem.


\bibliography{ref}

\clearpage 

\appendix
\section*{Appendix} %

\urlstyle{rm} 
\def\UrlFont{\rm}  

\frenchspacing  
\setlength{\pdfpagewidth}{8.5in} 
\setlength{\pdfpageheight}{11in} 
\setcounter{secnumdepth}{2}

\pdfinfo{
/TemplateVersion (2026.1)
}

\title{Appendix for CoCoNUTS: Concentrating on Content while Neglecting Uninformative Textual Styles for AI-Generated Peer Review Detection}

\appendix

\section{Detailed Definition of Content and Style}
\label{sec:appendix_definitions}

The conceptual distinction between content and style is foundational to our work. We posit that reliable AI-generated text detection must prioritize content over style—a premise that necessitates clear, operational definitions for these two concepts. These definitions are not merely theoretical; they form the architectural underpinning for our CoCoNUTS benchmark and the multi-task learning objective of our CoCoDet detector. They inform the principles behind our data construction and the mechanism by which our model isolates authentic signals from superficial artifacts.

\subsection{Content: The Semantic Core}

We define \textbf{content} as the invariant semantic core of a text—its underlying meaning, logical propositions, factual claims, and expressed ideas. Content addresses the question, “What is said?” It represents the informational payload that should, in principle, remain invariant across different phrasings or translations. In the context of peer review, this includes the reviewer's critical assessments, their summary of a paper's contributions, and specific, actionable recommendations.

Our central hypothesis is that while style is highly malleable, the provenance of complex, domain-specific ideas serves as a more reliable and difficult-to-forge signal of authorship. Therefore, our detection task is framed as a problem of content composition identification.

\subsection{Style: The Form of Expression}

Conversely, we define \textbf{style} as the set of formal properties governing how content is expressed. Style answers the question, “How is it being said?” and encompasses a wide spectrum of features that can be altered without fundamentally altering the core meaning. We categorize these features into a three-level hierarchy:

\begin{itemize}
    \item \textbf{Linguistic Features}: This level constitutes the foundational lexical choices (e.g., vocabulary, formality, synonym preference) and syntactic patterns (e.g., sentence length, complexity, use of active vs. passive voice).

    \item \textbf{Discourse-Level Features}: This higher level relates to the text's overall structure and rhetorical strategy. It includes the ordering of arguments, the use of logical transitions, and the persuasive or critical tone of the writing.

    \item \textbf{Statistical Artifacts}: Crucially for our task, this category covers the subtle, often subconscious statistical regularities that act as a statistical signature of the author—be it a human or a specific LLM. These include low-level patterns like perplexity (PPL) and n-gram frequency distributions, as well as more overt indicators like characteristic boilerplate phrases.
\end{itemize}

\section{Dataset Details}
\label{sec:appendix_a}

\subsection{Prompt for Dataset Construction}

We used the following prompts for review generation:

\begin{myprompt}{HW\&MT}
\label{prompt:hwmt}
\

\# EN2CN:

You are a professional AI field translator. Please translate the following English peer review into Chinese, paying attention to:

    1. Technical terms should be accurately translated
    
    2. Maintain the original review structure
    
    3. Keep the academic rigor while making it fluent in Chinese
    
    4. Output only the translated text, without any other information

Review:
\{en\_review\}

\# CN2EN:

You are a professional AI field translator. Please translate the following Chinese peer review into English, paying attention to:

    1. Technical terms should be accurately translated
    
    2. Maintain the original review structure
    
    3. Keep the academic rigor while making it fluent in English
    
    4. Output only the translated text, without any other information
    
Review:
\{cn\_review\}
\end{myprompt}

\begin{myprompt}{HW\&MP}
\label{prompt:hwmp}
\

You are a senior AI researcher and experienced reviewer for top-tier AI conferences.  Please polish the following peer review.
Please maintain the original technical content and core evaluation while improving sentence structure, terminology consistency, and readability.

Review:
\{review\}

Only output the polished review,do not include any other details.
\end{myprompt}

\begin{myprompt}{HW\&MG}
\label{prompt:hwmg}
\ 

You are a senior AI researcher and an experienced reviewer for top-tier AI conferences. Your task is to polish and expand a user-provided peer review.
Your goal is to elevate it into a high-quality, professional piece by:

1. Delete some redundant content make it more concise.

2. Expanding its content based on the provided paper content.

3. Improve the sentence structure, terminology consistency, and readability.

4. Output: Provide only the raw text of the elevated review , do not include any other details.

Review:
\{review\}

Paper content
: \{paper\_content\}
\end{myprompt}

\begin{myprompt}{MG}
\label{prompt:mg}
\ 

You are a senior AI researcher and experienced reviewer for top-tier AI conferences. Please carefully read the example reviews and then analyze the paper content provided by the user. After that write a comprehensive and objective review of the paper.
        Please follow the basic review content requirements(e.g., summary, evaluation, questions, suggestions for improvement) and ground your evaluation in the provided paper content.
        
Here are two examples of reviews:

**Example 1:**
\{example1\}

**Example 2:**
\{example2\}

Paper content
: \{paper\_content\}

Please only output the review, do not include any other details.

\end{myprompt}

\begin{myprompt}{MG\&MP}
\label{prompt:mgmp}
\ 

You are a senior AI researcher and experienced reviewer for top-tier AI conferences. 
Please paraphrase the review given by the user to make it more natural and human-written.

Review:
\{review\}

Only output the paraphrased review, do not include any other details.
\end{myprompt}

\subsection{Dataset Split}
\label{appendix:dataspilt}
We partitioned the complete CoCoNUTS dataset into training, validation, and test sets using an approximate 8:1:1 ratio. 

To ensure that each set is a representative sample of the overall data distribution, we employed a stratified sampling strategy. The stratification was performed based on the generating or modifying LLM for each instance. This approach guarantees that the data from every language model used in our construction process is proportionally represented across the training, validation, and test sets. 

This model-level stratification is crucial for preventing a model from being evaluated on a significantly different distribution of AI-generated styles than it was trained on. It ensures that our evaluation robustly measures the detector's ability to handle a consistent and diverse mix of AI sources. The detailed statistics for each split are presented in Table~\ref{tab:my_training_set_stats}, Table~\ref{tab:my_corrected_validation_set_stats}, and Table~\ref{tab:my_test_set_stats}.
\begin{table}[h!]
  \centering
  \small
  {
    \setlength{\tabcolsep}{2pt} 
    \begin{tabular}{@{}cccc@{}}
      \toprule
      \multicolumn{4}{l}{\textbf{Human} \hfill 84,104} \\
      \cmidrule(lr){1-1} \cmidrule(lr){2-3}
      \multicolumn{1}{c}{\textbf{HW}} & \multicolumn{2}{c}{\textbf{HWMT}} & \\
      Human & Llama & Qwen2.5 &  \\
      28,069 & 13,724 & 14,313 &  \\
      \cmidrule(lr){1-4}
      \multicolumn{4}{c}{\textbf{HWMP}} \\
      Gemini & Llama & Qwen3 & \\
      8,287 & 9,938 & 9,773 &  \\
      \midrule 

      \multicolumn{4}{l}{\textbf{Mix} \hfill 84,184} \\
      \cmidrule(lr){1-4}
      \multicolumn{4}{c}{\textbf{HWMG}} \\
      Gemini & Llama & Qwen2.5 & Qwen3 \\
      15,392 & 35,230 & 23,393 & 10,169 \\
      \midrule 

      \multicolumn{4}{l}{\textbf{AI} \hfill 84,139} \\
      \cmidrule(lr){1-4}
      \multicolumn{4}{c}{\textbf{MG}} \\
      Claude & DeepSeek & Gemini & GPT4o \\
      6,062  & 2,800  & 6,400 & 6,077 \\
      Llama & Qwen2.5 & Qwen3 &  \\
      10,800 & 8,000 & 8,000 &  \\
      \cmidrule(lr){1-4}
      \multicolumn{4}{c}{\textbf{MGMP}} \\
      Gemini-Llama & Llama-Gemini & Llama-Qwen2.5 & DS-Gemini  \\
      5,600 & 6,400 & 5,600 & 2,400 \\
      DS-Llama & Qwen2.5-Gemini & Qwen3-Gemini & Qwen3-Llama \\
      4,000 & 3,200 & 4,000  & 4,800 \\
      \bottomrule
    \end{tabular}%
  } 
    \caption{Statistics of our training set.}
    \label{tab:my_training_set_stats}
\end{table}%

\begin{table}[h!]
  \centering
  \small
  {
    \setlength{\tabcolsep}{2pt} 
    \begin{tabular}{@{}cccc@{}}
      \toprule
      \multicolumn{4}{l}{\textbf{Human} \hfill 10,553} \\
      \cmidrule(lr){1-1} \cmidrule(lr){2-3}
      \multicolumn{1}{c}{\textbf{HW}} & \multicolumn{2}{c}{\textbf{HWMT}} & \\
      Human & Llama & Qwen2.5 &  \\
      3,495 & 1,703 & 1,815 &  \\
      \cmidrule(lr){1-4}
      \multicolumn{4}{c}{\textbf{HWMP}} \\
      Gemini & Llama & Qwen3 & \\
      1,050 & 1,264 & 1,226 &  \\
      \midrule 

      \multicolumn{4}{l}{\textbf{Mix} \hfill 10,483} \\
      \cmidrule(lr){1-4}
      \multicolumn{4}{c}{\textbf{HWMG}} \\
      Gemini & Llama & Qwen2.5 & Qwen3 \\
      1,923 & 4,379 & 2,905 & 1,276 \\
      \midrule 

      \multicolumn{4}{l}{\textbf{AI} \hfill 10,518} \\
      \cmidrule(lr){1-4}
      \multicolumn{4}{c}{\textbf{MG}} \\
      Claude & DeepSeek & Gemini & GPT4o \\
      758  & 350  & 800 & 760 \\
      Llama & Qwen2.5 & Qwen3 &  \\
      1,350 & 1,000 & 1,000 &  \\
      \cmidrule(lr){1-4}
      \multicolumn{4}{c}{\textbf{MGMP}} \\
      Gemini-Llama & Llama-Gemini & Llama-Qwen2.5 & DS-Gemini  \\
      700 & 800 & 700 & 300 \\
      DS-Llama & Qwen2.5-Gemini & Qwen3-Gemini & Qwen3-Llama \\
      500 & 400 & 500  & 600 \\
      \bottomrule
    \end{tabular}%
  } 
    \caption{Statistics of our corrected validation set.}
    \label{tab:my_corrected_validation_set_stats}
\end{table}%

\begin{table}[h!]
  \centering
  \small
  {
    \setlength{\tabcolsep}{2pt} 
    \begin{tabular}{@{}cccc@{}}
      \toprule
      \multicolumn{4}{l}{\textbf{Human} \hfill 10,523} \\
      \cmidrule(lr){1-1} \cmidrule(lr){2-3}
      \multicolumn{1}{c}{\textbf{HW}} & \multicolumn{2}{c}{\textbf{HWMT}} & \\
      Human & Llama & Qwen2.5 &  \\
      3,496 & 1,715 & 1,790 &  \\
      \cmidrule(lr){1-4}
      \multicolumn{4}{c}{\textbf{HWMP}} \\
      Gemini & Llama & Qwen3 & \\
      1,035 & 1,248 & 1,239 &  \\
      \midrule 

      \multicolumn{4}{l}{\textbf{Mix} \hfill 10,513} \\
      \cmidrule(lr){1-4}
      \multicolumn{4}{c}{\textbf{HWMG}} \\
      Gemini & Llama & Qwen2.5 & Qwen3 \\
      1,936 & 4,388 & 2,903 & 1,286 \\
      \midrule 

      \multicolumn{4}{l}{\textbf{AI} \hfill 10,518} \\
      \cmidrule(lr){1-4}
      \multicolumn{4}{c}{\textbf{MG}} \\
      Claude & DeepSeek & Gemini & GPT4o \\
      758  & 350  & 800 & 760 \\
      Llama & Qwen2.5 & Qwen3 &  \\
      1,350 & 1,000 & 1,000 &  \\
      \cmidrule(lr){1-4}
      \multicolumn{4}{c}{\textbf{MGMP}} \\
      Gemini-Llama & Llama-Gemini & Llama-Qwen2.5 & DS-Gemini  \\
      700 & 800 & 700 & 300 \\
      DS-Llama & Qwen2.5-Gemini & Qwen3-Gemini & Qwen3-Llama \\
      500 & 400 & 500  & 600 \\
      \bottomrule
    \end{tabular}%
  } 
    \caption{Statistics of our test set.}
    \label{tab:my_test_set_stats}
\end{table}%

\subsection{Detailed Dataset Statistics}
To provide a comprehensive statistical overview of our CoCoNUTS dataset, we present a detailed analysis of its textual properties, focusing on length and distribution. These statistics underscore the distinct characteristics of the six fine-grained modes and the three content-based classes, which are central to our benchmark's design. This analysis reveals key differences in text length and structure across categories, which may influence detector performance and highlight the challenges of distinguishing between different forms of human and AI collaboration.
\begin{figure}[h!]
    \centering
    \includegraphics[width=0.95\linewidth]{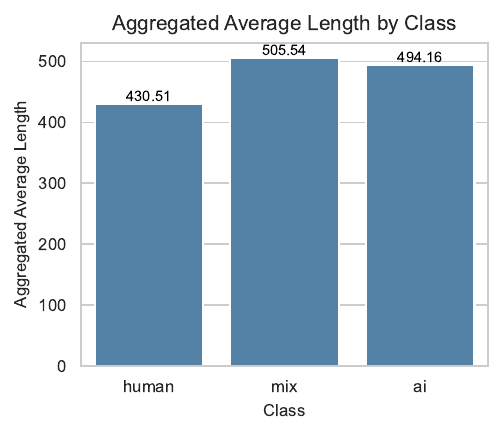}
    \caption{Aggregated average length (in words) across the three content-composition classes: Human, Mix, and AI. Peer reviews with substantive content generated by AI demonstrates longer length. }
    \label{fig:class_avg_length}
\end{figure}

\begin{figure}[h!]
    \centering
    \includegraphics[width=0.95\linewidth]{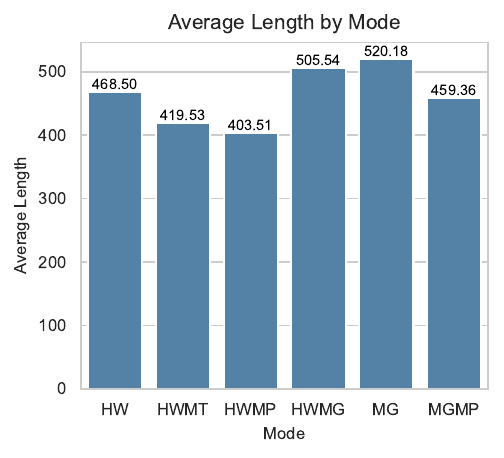}
    \caption{Average text length (in words) for each of the six collaboration modes. Modes involving direct machine generation (MG, HWMG) tend to be longer, while post-processing steps like polishing (HWMP) or paraphrasing (MGMP) can result in shorter texts compared to their respective source modes.}
    \label{fig:mode_avg_length}
\end{figure}

We present a statistical analysis of the average text length across different classes in our CoCoNUTS dataset, revealing systematic variations based on the nature of AI involvement. As illustrated in Figure~\ref{fig:class_avg_length}, reviews with substantive AI contributions exhibit a  greater length than purely human-authored texts. Specifically, the \texttt{Human} class has an average length of approximately 431 words, whereas the \texttt{Mix} and \texttt{AI} classes are substantially longer, at 506 and 494 words, respectively. This suggests that the process of content generation by large language models, whether partial or complete, tends to produce more verbose outputs compared to the human baseline.
\begin{figure*}[t!]
    \centering
    \begin{subfigure}[b]{0.32\linewidth}
        \includegraphics[width=\linewidth]{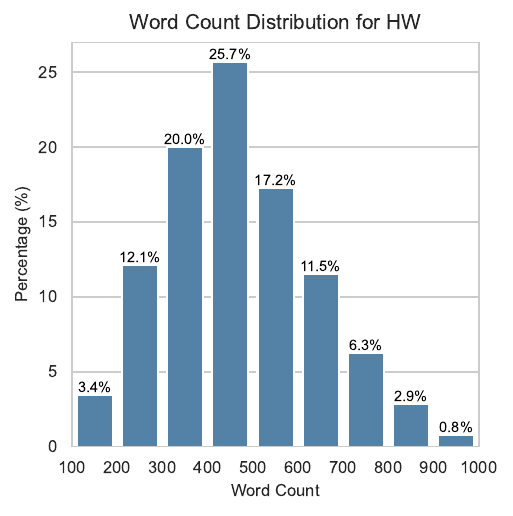}
        \subcaption{HW Distribution.}
        \label{fig:dist_hw}
    \end{subfigure}
    \hfill
    \begin{subfigure}[b]{0.32\linewidth}
        \includegraphics[width=\linewidth]{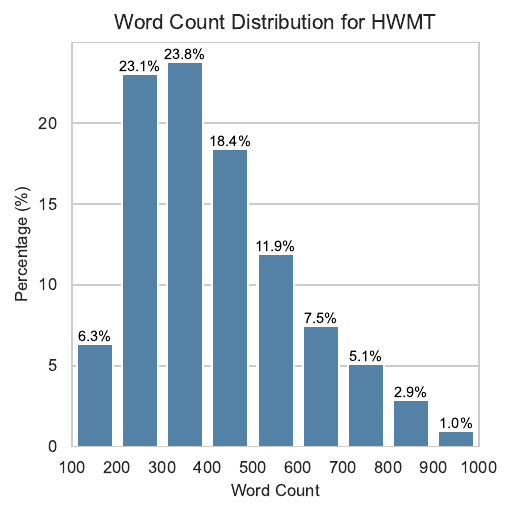}
        \subcaption{HWMT Distribution.}
        \label{fig:dist_hwmt}
    \end{subfigure}
    \hfill
    \begin{subfigure}[b]{0.32\linewidth}
        \includegraphics[width=\linewidth]{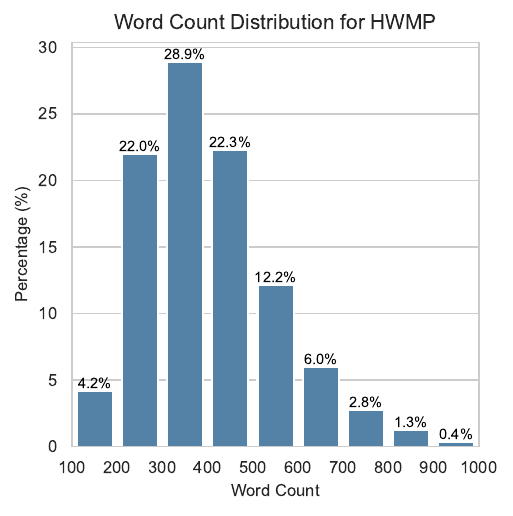}
        \subcaption{HWMP Distribution.}
        \label{fig:dist_hwmp}
    \end{subfigure}
    
    \vspace{1em} 

    \begin{subfigure}[b]{0.32\linewidth}
        \includegraphics[width=\linewidth]{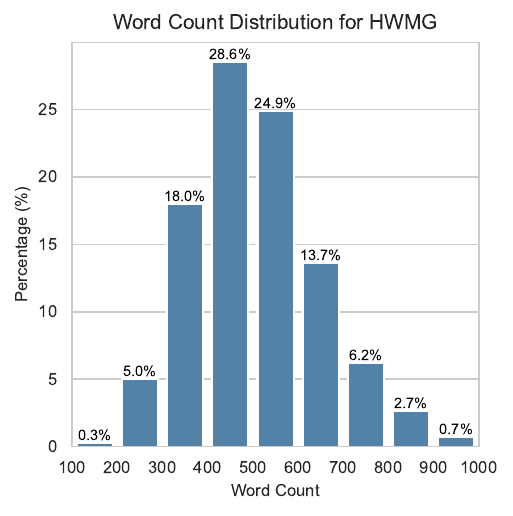}
        \subcaption{HWMG Distribution.}
        \label{fig:dist_hwmg}
    \end{subfigure}
    \hfill
    \begin{subfigure}[b]{0.32\linewidth}
        \includegraphics[width=\linewidth]{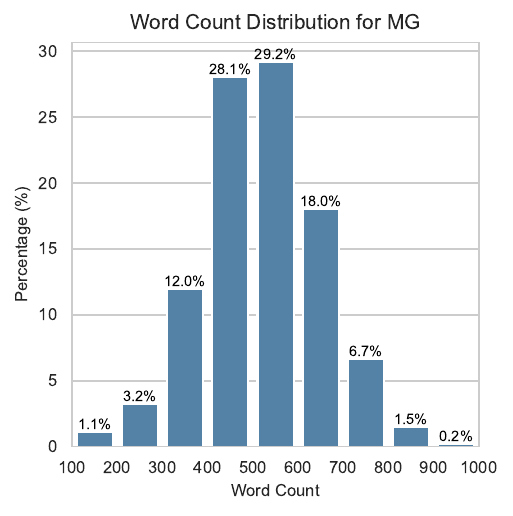}
        \subcaption{MG Distribution.}
        \label{fig:dist_mg}
    \end{subfigure}
    \hfill
    \begin{subfigure}[b]{0.32\linewidth}
        \includegraphics[width=\linewidth]{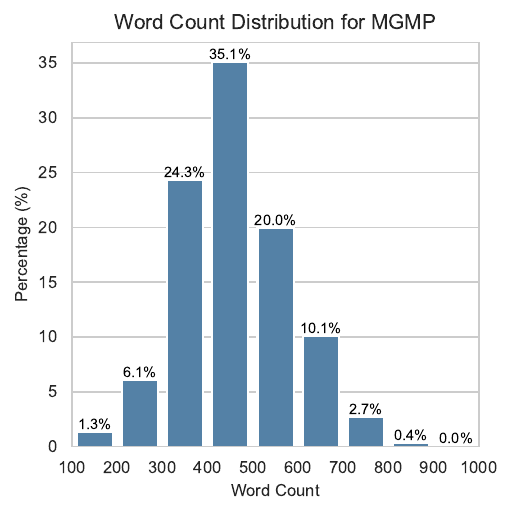}
        \subcaption{MGMP Distribution.}
        \label{fig:dist_mgmp}
    \end{subfigure}

    \caption{Word count distributions for each of the six fine-grained collaboration modes in the CoCoNUTS dataset. This detailed view reveals the distinct length characteristics of each mode, such as the longer tails for machine-generated content (e.g., MG, HWMG) and the condensing effect of polishing (HWMP).}
    \label{fig:word_count_distributions}
\end{figure*}

A more granular analysis, presented in Figure~\ref{fig:mode_avg_length}, clarifies the nuanced effects of different human-AI collaboration modes. The purely machine-generated mode produces the longest texts, averaging 520 words. However, a critical observation is that the semantic-invariant operations reduce text length. For instance, when human-written reviews (469 words) are subjected to machine polishing or machine translation, their average lengths decrease to 404 and 420 words, respectively. A similar condensing effect is observed when machine-generated texts are paraphrased, reducing the average length from 520 to 459 words.

\section{Experimental Setup  Details}
\label{appendix:experimentsetup}

\subsection{General Detectors Details}

\paragraph{LLMDet:}LLMDet provides a method for model-specific detection of AI-generated text without requiring real-time access to the source LLMs during inference. The approach operates in two phases. First, an offline 'dictionary construction' phase builds a fingerprint for each target LLM. This involves pre-computing and storing frequent n-gram patterns and their corresponding next-token probability distributions. During the online detection phase, LLMDet calculates a 'proxy perplexity' score for the input text against each pre-computed dictionary. These proxy scores are then used as features for a classifier to identify the most likely source model.

\paragraph{RADAR:}RADAR  is designed to address a critical vulnerability in AI text detectors: their susceptibility to paraphrasing attacks. It employs an adversarial training framework where a detector model and a paraphraser model are trained in opposition. The paraphraser's objective is to rewrite AI-generated text to evade detection, while the detector is simultaneously trained to correctly identify not only original AI text but also these adversarially paraphrased versions. This process forces the detector to learn more robust features that are resilient to stylistic modifications, thereby improving its performance in realistic, adversarial scenarios.

\paragraph{Fast-DetectGPT:}Fast-DetectGPT presents a zero-shot detection method that significantly improves both the efficiency and accuracy of identifying machine-generated text. The core of the method is a new metric called `conditional probability curvature.' This metric is based on the hypothesis that machine-generated text exhibits a different statistical pattern in its conditional probability distribution compared to human-written text. Instead of re-evaluating multiple perturbed versions of a passage, Fast-DetectGPT computes this curvature by analyzing alternative token probabilities obtained from a single forward pass of a scoring model, drastically reducing computational overhead.

\paragraph{Binoculars:} Binoculars introduces a zero-shot detection method that identifies AI-generated text by analyzing it from two perspectives using two closely related language models: an observer and a performer. The core idea is to compute a ratio between the text's standard perplexity (as seen by the observer) and its cross-perplexity, which measures the predictive divergence between the two models. This ratio creates a robust statistical signature that effectively distinguishes machine-generated text, which shows high model-to-model agreement, from human-written text, which exhibits greater variability. The method is notably effective at normalizing for unusual prompts that can otherwise mislead simpler perplexity-based detectors.

\paragraph{LLM-DetectAIve:}LLM-DetectAIve advances beyond binary classification by providing a fine-grained detection system. Instead of simply labeling text as human or machine, it distinguishes between four categories: purely human-written, purely machine-generated, machine-generated text that has been humanized, and human-written text that has been ``polished'' by a machine. This nuanced approach is designed to differentiate between acceptable uses of LLMs (e.g., polishing) and attempts to obfuscate AI authorship, making it particularly relevant for educational and academic contexts.
\subsection{Prompts for LLM-based Detectors}
\label{appendix:prompt4llmdetector}
The exact prompts for LLM-based Detectors of both zero-shot and few-shot settings are shown below:
\begin{myprompt}{Zero-shot}
\label{Zero-shot}
\ 

 You are an expert AI generated peer review detector. Your task is to classify the given text into one of three categories based on the content: 
        Follow these rules precisely:
        
        1.  **`human`**: Classify as `human` if the core content was written by a human. 
        This includes texts that were later machine-translated or polished by language tools.
        
        2.  **`ai`**: Classify as `ai` if the core content was generated by an AI. 
        This includes texts that were later edited or 'humanized' by a person to sound more natural.
        
        3.  **`mix`**: Classify as `mix` only if the text contains substantive content contributions from both human and AI. This includes some sections written by a human and others generated by an AI.
        
        Your response must be *only* one of these three words. Do not provide any explanations or additional text.

\end{myprompt}

\begin{myprompt}{Few-shot}
\label{Few-shot}
\ 

 You are an expert AI generated peer review detector. Your task is to classify the given text into one of three categories based on the content: 
        Follow these rules precisely:
        
        1.  **`human`**: Classify as `human` if the core content was written by a human. 
        This includes texts that were later machine-translated or polished by language tools.
        
        2.  **`ai`**: Classify as `ai` if the core content was generated by an AI. 
        This includes texts that were later edited or 'humanized' by a person to sound more natural.
        
        3.  **`mix`**: Classify as `mix` only if the text contains substantive content contributions from both human and AI. This includes some sections written by a human and others generated by an AI.
        
        Your response must be *only* one of these three words. Do not provide any explanations or additional text.

**Example for `human`:**

``human\_example''

Correct Answer: human

**Example for `mix`:**

``mix\_example''

Correct Answer: mix

**Example for `ai`:**

``ai\_example''

Correct Answer: ai
\end{myprompt}

\subsection{Training Hyperparameters}
\label{appendix:traininghyper}
The final hyperparameters used to fine-tune our CoCoDet model, determined through a sequential grid search on the validation set, are detailed in Table \ref{tab:hyperparams}.
\begin{table}[h!]
  \centering
  \begin{tabular}{ll}
    \toprule
    \textbf{Parameter} & \textbf{Value} \\
    \midrule
    Epochs               & 5 \\
    Max Sequence Length  & 2048 \\
    Learning Rate        & 2e-5 \\
    Batch Size           & 16 \\
    Weight Decay         & 0.01 \\
    Auxiliary Weights ($\alpha, \beta, \gamma$) & 0.4, 0.4, 0.2 \\
    Base Margin       & 0.25 \\
    Cost Margin          & 0.25 \\
    Scaling Factor    & 30 \\
    \bottomrule
  \end{tabular}
  \caption{Final hyperparameters for fine-tuning the CoCoDet model. These values were selected based on the best performance on the validation set after a sequential grid search.}
  \label{tab:hyperparams}
\end{table}

\section{Additional Analysis}
\subsection{Analysis of Content Consistency After Semantic-Invariant Operations}
\label{appendix:semanticinv}
This part provides case studies (Tables~\ref{tab:case_hwmt}, \ref{tab:case_hwmp}, and \ref{tab:case_mgmp}) to demonstrate that semantic-invariant operations like machine translation (MT) and polishing (MP) alter only the stylistic presentation of a peer review, not its core content. These examples support our data construction methodology, where such operations are treated as stylistic modifications.


\begin{table*}[bthp]
    \centering

    \begin{tabularx}{\linewidth}{XX}
    \toprule
    \textbf{HW} & \textbf{HW \& MT} \\
    \midrule
    This paper first proves that inter-class distinctiveness and intra-class compactness among hash codes determine the lower bound of hash codes'  performance. And it shows that promoting these two characteristics could lift the bound and improve hash learning. Then it proposes a surrogate model to fully exploit such objective by estimating posterior of hash codes. Extensive experiments reveal effectiveness of the proposed method.
    
    Strengths:
    
    1. The studied problem is interesting and important because a theoretical analysis on criteria of learning good hash codes remains largely unexplored.
    
    2. The proposed method seems to be reasonable and effective.
    
    3. Experiment seems to be extensive. 
    
    Weaknesses:
    
    1. There exist some typos and grammatical errors in the paper. 
    
    2. The training sets on all datasets are relatively small.
    
    Is the performance improvement still significant when we increase the size of training set? 
    
    The authors have adequately addressed the limitations and potential negative societal impact of their work.
    & 
    This paper first demonstrates that the inter-class distinguishability and intra-class compactness determine the lower bound of hash code performance. It also shows that promoting these two properties can improve the lower bound and enhance hash learning. Then, the paper proposes a proxy model that leverages this objective by estimating the posterior probability of hash codes. Extensive experiments reveal the effectiveness of the proposed method.
    
    Strengths:
    
    1. The research question is interesting and important, as theoretical analysis of the criteria for learning good hash codes has not been sufficiently explored.
    
    2. The proposed method appears reasonable and effective.
    
    3. The experiments seem quite extensive.
    
    Weaknesses:
    
    1. There are some spelling and grammatical errors in the paper.
    
    2. The training sets for all datasets are relatively small.
    
    Does the performance improvement remain significant when the size of the training set is increased?
    
    The authors adequately address the limitations and potential negative societal impacts of their work.
    \\
    \bottomrule
    \end{tabularx}
    \caption{Content Consistency in Machine Translation (HW \& MT)}
    \label{tab:case_hwmt}
\end{table*}


\begin{table*}[htbp]
    \centering

    \begin{tabularx}{\linewidth}{XX}
    \toprule
    \textbf{HW} & \textbf{HW \& MP} \\
    \midrule
    This paper proposed a new black-box attack method in the hard-label setting. By using a well-designed policy network in a novel reinforcement learning formulation, the new method learns promising search directions of the adversarial examples and showed that query complexity is significantly reduced in experiments.

    A couple questions I have:

    How are $\beta_1$ and $\beta_2$ chosen?

    Is there any convergence analysis? How do we guarantee the proposed attacking method will converge?
    
    The proposed method is introduced as an attack that minimize $L_2$ distance.  Is it possible to extend this attacking method to $L_{inf}$?
    
    I am curious, have you tried evaluate the attacking methods on DenseNet as the victim model?

    Overall, I think this paper is very readable and is clearly written with a very good background and context. 
    I found the idea of the paper original and interesting. And the authors have conducted experiments that show their new method has the best query efficiency, which is reasonable and aligns with their idea.  For cons, this paper does not have a convergence analysis. And if the experiments could be conducted on more data sets and more victim models, then it would be more convincing.

    & 
    This paper introduces a novel black-box attack method operating in the hard-label setting. The core contribution is a reinforcement learning formulation that employs a meticulously designed policy network to learn promising search directions for adversarial examples. This approach demonstrably reduces query complexity, as evidenced by experimental results.

    I have a few questions regarding the method:

    1. How are the hyperparameters $\beta_1$ and $\beta_2$ determined?

    2. Is there any convergence analysis for the proposed attack method? How can we guarantee its convergence?

    3. The proposed method is presented as an L2-norm-minimizing attack. Is it possible to extend this approach to the L-infinity norm setting?
    
    4. Have the authors evaluated the attack method against DenseNet as a victim model?

    Overall, the paper is well-written, clear, and provides excellent background and context. The core idea is original and compelling. The authors have conducted experiments that effectively demonstrate the superior query efficiency of their new method, which aligns well with the proposed concept. On the downside, the paper lacks a convergence analysis. Furthermore, the experimental results would be more convincing if evaluated on a broader range of datasets and victim models.
    \\
    \bottomrule
    \end{tabularx}
    \caption{Content Consistency in Machine Polishing (HW \& MP)}
    \label{tab:case_hwmp}
\end{table*}



\begin{table*}[t]
    \centering

    \begin{tabularx}{\linewidth}{XX}
    \toprule
    \textbf{MG} & \textbf{MG \& MP} \\
    \midrule
    The paper addresses the gradient mismatch problem in training Quantized Neural Networks (QNNs) via a novel approach called ManiQuant, which leverages Riemannian manifold learning to revise the Straight-Through Estimator (STE). The core idea is to model the parameter space as a metric tensor in a Riemannian manifold, enabling a more accurate gradient approximation through the non-differentiable quantization function. This is a theoretically intriguing direction, as it introduces geometric priors into gradient estimation, which has been relatively underexplored in the QNN literature. The experimental results on CIFAR10/100 and ImageNet suggest improved stability and performance compared to standard STE, supported by ablation studies. However, the review identifies several key issues. First, the technical details of how the Riemannian metric tensor is defined and computed remain under-specified. For instance, it is unclear whether the metric is learned during training or derived from a fixed geometric structure (e.g., based on curvature or data distribution). This lack of clarity weakens the reproducibility and interpretability of the method. Second, the paper primarily benchmarks against STE-based methods but does not compare with non-STE alternatives like differentiable quantization layers or optimization-based quantization frameworks. A more comprehensive comparison with existing QNN strategies would strengthen the evaluation. Third, while the results on standard image datasets are promising, extending experiments to more challenging tasks (e.g., video, NLP) or larger-scale models (e.g., ViT, ResNet-152) would better demonstrate the method's generalizability. Finally, the theoretical justification for why Riemannian embeddings reduce gradient mismatch is not fully developed. A derivation linking the manifold structure to gradient stability would enhance the contribution. Overall, this work offers a novel geometric perspective on STE but requires deeper technical exposition and broader empirical validation to establish its significance for the QNN community.
    & 
    This paper proposes a fresh approach to addressing the gradient mismatch issue in Quantized Neural Networks (QNNs) through ManiQuant, a method that utilizes Riemannian manifold learning to refine the Straight-Through Estimator (STE). By treating the parameter space as a Riemannian manifold with a metric tensor, ManiQuant achieves a more precise gradient approximation, even in the face of non-differentiable quantization functions. This geometric perspective on gradient estimation is a compelling, albeit underexplored, direction in QNN research. Experimental results on benchmark datasets such as CIFAR10/100 and ImageNet indicate that ManiQuant offers improved stability and performance over traditional STE methods, with supporting evidence from ablation studies. However, several concerns need to be addressed. The definition and computation of the Riemannian metric tensor, for instance, are not thoroughly explained, leaving unclear whether it's learned during training or predetermined by a specific geometric structure. This omission hinders the method's reproducibility and interpretability. Furthermore, the comparison to other quantization methods is limited, primarily focusing on STE-based approaches while neglecting alternative strategies such as differentiable quantization layers or optimization-based frameworks. A more inclusive comparison would significantly enhance the evaluation. Additionally, while the current results are promising, they are based on standard image datasets, and extending the experiments to more complex tasks or larger models would provide a more robust demonstration of the method's versatility. Lastly, a detailed theoretical explanation of how Riemannian embeddings mitigate gradient mismatch is lacking. Developing this justification would substantially bolster the paper's contribution. In essence, ManiQuant introduces an innovative geometric viewpoint to STE, but to firmly establish its impact on the QNN community, it requires more comprehensive technical details and empirical validations.

    \\
    \bottomrule
    \end{tabularx}
    \caption{Content Consistency in Machine Polishing (MG \& MP)}
    \label{tab:case_mgmp}
\end{table*}

\subsection{Analysis of Errors in LLM Reasoning}
\label{appendix:erroranalysis}
To investigate the failure modes of Large Language Models (LLMs) as detectors, we examined their Chain-of-Thought (CoT) reasoning. Our analysis reveals a consistent limitation: LLMs tend to perform a surface structural assessment of the text rather than a deep analysis of its substantive content.

The reasoning processes of Qwen3 and DeepSeek, shown below, exemplify this behavior. Both models begin by correctly identifying the document's structure and topic, for instance, recognizing a review's ``Strengths'' and ``Weaknesses'' sections or its subject matter. This initial step resembles a form of content analysis.

However, the models' subsequent reasoning for their final classification pivots to stylistic and qualitative evaluations. They base their decision on perceived attributes such as ``technical language'', ``nuanced points'', or ``natural flow''. In essence, the models equate the successful imitation of an expert's writing style and structure with genuine human authorship. They do not demonstrate an ability to question the provenance of the well-formed arguments themselves.

This finding suggests that prompt-based LLM detectors primarily operate by matching text to a learned schema of what an expert review should look like, rather than by performing true content-based source attribution. This reliance on form and style, rather than substance, makes them unreliable for distinguishing high-quality AI-generated text from human writing. 

\begin{mdframed}[style=ExampleBox, frametitle={An Error Case of the CoT of Qwen3}]
Okay, let's tackle this. The user wants me to determine if the given text is human, AI, or a mix. The examples provided show that ``human'' is when the core is human-written, even if edited by AI. ``AI'' is when the core is AI-generated, even if a human polished it. ``Mix'' is when content is from human and AI.

Looking at the text in question: it's a detailed review of a paper on reward imputation in contextual batched bandits. The review discusses the method, analysis, experiments, and suggests improvements. The language is technical, with specific references to theoretical guarantees, experimental results, and comparisons. 

The structure is formal and thorough, which is typical in academic reviews. However, AI-generated reviews might have a more polished or generic tone, but this one has specific critiques like suggesting comparisons with other techniques, hyperparameter sensitivity, and applicability to other settings. These are nuanced points that a human reviewer might note based on their expertise. 

Also, the text includes phrases like ``the paper is well-crafted'' and ``the claims seem accurate'', which are subjective judgments a human might make. AI might generate such a review, but the depth and specificity here lean more towards a human author. There's no clear indication of distinct AI-generated sections mixed with human parts. 

Therefore, I think this is a human-written review. The detailed, critical analysis and specific suggestions are hallmarks of a human reviewer's input.
\end{mdframed}

\begin{mdframed}[style=ExampleBox, frametitle={An Error Case of the CoT of DeepSeek-R1}]
Alright, I need to classify this text as `human', `ai', or `mix' based on the given rules. Let me start by reading the content carefully. 

The text is a review of a research paper submission. It discusses strengths, weaknesses, and suggestions for improvement. The structure is organized with clear headings like ``Strengths'', ``Weaknesses'', and ``Questions and Suggestions for Improvement''. The language is critical but constructive, pointing out areas where the paper lacks detail, such as empirical evaluations and mathematical formulations. 

Looking at the examples provided, the 'human' example was detailed with specific critiques and personal observations. The 'ai' example was more structured and used bullet points but still had a formal tone. The 'mix' example had sections where parts seemed human-written and others AI-generated.

Here, the reviewer uses phrases like ``The paper could benefit from'' and ``The authors should consider...'' which suggest personal judgment. The questions raised are specific, like handling high-dimensional settings, which indicates deep understanding. The suggestions for including empirical evaluations and clearer structure show critical thinking typical of a human reviewer. There's no obvious sign of AI-generated content, like overly formal or generic statements. The entire text flows naturally without sections that stand out as AI-generated. 
\end{mdframed}

\end{document}